\documentclass[10pt, a4paper, onecolumn]{article}

\usepackage{arxiv}
\usepackage[toc]{appendix}
\usepackage[utf8]{inputenc} % allow utf-8 input
\usepackage[T1]{fontenc}    % use 8-bit T1 fonts
\usepackage{url}            % simple URL typesetting
\usepackage{booktabs}       % professional-quality tables
\usepackage{amsfonts}       % blackboard math symbols
\usepackage{nicefrac}       % compact symbols for 1/2, etc.
\usepackage{microtype}      % microtypography
\usepackage{lipsum}
\usepackage{amsmath}
\usepackage{cite}
\usepackage{amssymb}
\usepackage{verbatim}
\usepackage{mdwlist} % for compact lists
\usepackage{color}

\usepackage{algpseudocode}
\usepackage{algorithmicx}
\usepackage[linesnumbered,ruled]{algorithm2e}
\usepackage{lineno}
\usepackage{framed,multirow}
\usepackage[final]{pdfpages}
\usepackage{latexsym}
\usepackage{wrapfig}

\usepackage{url}
\usepackage[table]{xcolor}
\usepackage{xcolor}
\usepackage{graphicx}
\usepackage{subfigure}
\usepackage{verbatim}
\usepackage{bm}
\usepackage{authblk}
\usepackage{tikz}
\usepackage{amsmath}
\usepackage[linesnumbered,ruled]{algorithm2e}
\usepackage[space]{grffile}
\usetikzlibrary{shapes,arrows}

\usepackage{graphicx}
\usepackage{amsmath}
\usepackage{amssymb}
\usepackage{booktabs}
\usepackage{subfigure}

\usepackage[pagebackref,breaklinks,colorlinks]{hyperref}

\usepackage[capitalize]{cleveref}
\crefname{section}{Sec.}{Secs.}
\Crefname{section}{Section}{Sections}
\Crefname{table}{Table}{Tables}
\crefname{table}{Tab.}{Tabs.}

\newcommand{\PE}{\mathrm{PE}}
\newcommand{\etal}{\textit{et al.}}
\newcommand{\eg}{\textit{e.g. }}

\title{RADAM: Texture Recognition through Randomized Aggregated Encoding of Deep Activation Maps}

\author[1,2]{Leonardo Scabini}
\author[1]{Kallil M. Zielinski}
\author[3]{Lucas C. Ribas}
\author[4]{Wesley N. Gonçalves}
\author[2]{Bernard De Baets}
\author[1]{Odemir M. Bruno}

\affil[1]{\small{S\~{a}o Carlos Institute of Physics, University of S\~{a}o Paulo, postal code 13560-970, São Carlos - SP, Brazil}}% (email: $\{$scabini,bruno$\}$@ifsc.usp.br)}}
\affil[2]{\small{KERMIT, Department of Data Analysis and Mathematical Modelling, Ghent University, Coupure links 653, postal code 9000, Ghent, Belgium}}% (email: $\{$Leonardo.Scabini,Bernard.DeBaets$\}$@UGent.be)}}
\affil[3]{\small{Institute of Biosciences, Humanities and Exact Sciences, São Paulo State University, postal code 15054-000, São José do Rio Preto - SP, Brazil}}% (email: lucas.ribas@unesp.br)}}
\affil[4]{\small{Faculty of Computing, Federal University of Mato Grosso do Sul, postal code 79070-900, Campo Grande - MS, Brazil}}% (email: wesley.goncalves@ufms.br)}}

\begin{document}
\maketitle
%%%%%%%%% ABSTRACT
\begin{abstract}
    Texture analysis is a classical yet challenging task in computer vision for which deep neural networks are actively being applied. Most approaches are based on building feature aggregation modules around a pre-trained backbone and then fine-tuning the new architecture on specific texture recognition tasks. Here we propose a new method named \textbf{R}andom encoding of \textbf{A}ggregated \textbf{D}eep \textbf{A}ctivation \textbf{M}aps (RADAM) which extracts rich texture representations without ever changing the backbone. The technique consists of encoding the output at different depths of a pre-trained deep convolutional network using a Randomized Autoencoder (RAE). The RAE is trained locally to each image using a closed-form solution, and its decoder weights are used to compose a 1-dimensional texture representation that is fed into a linear SVM. This means that no fine-tuning or backpropagation is needed. We explore RADAM on several texture benchmarks and achieve state-of-the-art results with different computational budgets. Our results suggest that pre-trained backbones may not require additional fine-tuning for texture recognition if their learned representations are better encoded.    
 %better pre-training of backbones significantly improves performance. %These findings suggest that the backbone is highly important for feature extraction, containing rich features 
 \end{abstract}

%%%%%%%%% BODY TEXT

\section{Introduction}\label{sec:intro}

For several decades, texture has been studied in Computer Vision as a fundamental visual cue for image recognition in several applications. Despite lacking a widely accepted theoretical definition, we 
all have developed an intuition for textures by analyzing the world around us from material surfaces in our daily life, through microscopic images, and even through macroscopic images from telescopes and remote sensing. In digital images, one abstract definition is that texture elements emerge from the local intensity constancy and/or variations of pixels producing spatial patterns roughly independently at different scales~\cite{scabini2020spatio}.

The classical approaches to texture recognition focus on the mathematical description of the textural patterns, considering properties such as statistics~\cite{julesz1962visual,haralick1979statistical,ojala2002multiresolution}, frequency~\cite{azencott1997texture,hoang2005color}, complexity/fractality~\cite{backes2012color,scabini2019multilayer}, and others~\cite{zhang2002brief}. Many such aspects of texture are challenging to model even in controlled imaging scenarios. Moreover, the wild nature of digital images also results in additional variability, making the task even more complex in real-world applications.

Recently, the power of deep neural networks has been extended to texture analysis by taking advantage of models pre-trained on big natural image datasets~\cite{FV-CNN,FV,DeepTEN,DEPNet,zhai2019,zhai2020,xue2020,condori2021,chen2021,yang2022}. These transfer-learning approaches combine the general vision capabilities of pre-trained models with dedicated techniques to capture additional texture information, achieving state-of-the-art performance on several texture recognition tasks. Therefore, most of the recent works on deep texture recognition propose to build new modules around a pre-trained deep network (backbone) and to retrain the new architecture for a specific texture analysis task. However, even if the new modules are relatively cheap in terms of computational complexity, resulting in good inference efficiency, the retraining of the backbone itself is usually costly. Going in a different direction, Randomized Neural Networks~\cite{schmidt1992feedforward,pao1992functional,pao1994learning,huang2006extreme} proposes a closed-form solution for training neural networks, instead of the common backpropagation, with various potential applications. For instance, the training time of randomization-based models was analyzed~\cite{RNN-AE} on datasets such as MNIST, resulting in gains up to 150 times. These gains can be expressive when hundreds of thousands of images are used to train a model.

%\subsection{Contributions}
% a section cannot have a single subsection; moreover, in case of multiple subsections, the first has to start after max one paragraph
In this work, we propose a new module for texture feature extraction from pre-trained deep convolutional neural networks (DCNNs). The method, called \textbf{R}andom encoding of \textbf{A}ggregated \textbf{D}eep \textbf{A}ctivation \textbf{M}aps (RADAM), goes in a different direction than recent literature on deep texture recognition. Instead of increasing the complexity of the backbone and then retraining everything, we propose a simple codification of the backbone features using a new randomized module. The method is based on aggregating deep activation maps from different depths of a pre-trained convolutional network, and then training Randomized Autoencoders (RAEs) in a pixel-wise fashion for each image, using a closed-form solution. This module outputs the decoder weights from the learned RAEs, which are used as a 1-dimensional feature representation of the input image. This approach is simple and does not require hyperparameter tuning or backpropagation training. Instead, we propose to attach a linear SVM at the top of our features, which can be simply used with standard parameters. Our code is open and is available in a public repository\footnote{\url{https://github.com/scabini/RADAM}}\label{github}. In summary, our main contributions are:

\begin{itemize}
    \item[(i)] We propose the RADAM texture feature encoding technique applied over a pre-trained DCNN backbone and coupled with a simple linear SVM. The model achieves impressive classification performance without needing to fine-tune the backbone, in contrast to what has been proposed in previous works.
    
    % \item[(ii)] Besides well-known backbones, we also explore RADAM over the MobileNet V2 and ConvNeXt architectures, achieving state-of-the-art performance with different computational budgets on several texture analysis benchmarks.
    
    \item[(ii)] Bigger backbones and better pre-training improve the performance of RADAM considerably, suggesting that our approach scales well.   
\end{itemize}

\section{Background}
We start by conducting a literature review on texture analysis with deep learning and randomized neural networks. The methods covered here are also considered for comparison in our experiments.

\subsection{Texture Analysis with Deep Neural Networks}

In this work, we focus on transfer-learning-based texture analysis by taking advantage of pre-trained deep neural networks. For a more comprehensive review of different approaches to texture analysis, the reader may consult~\cite{Liu2018}. There have been numerous studies involving deep learning for texture recognition, and here we review them according to two approaches: feature extraction or end-to-end fine-tuning. Some studies explore CNNs only for texture feature extraction and use a dedicated classifier apart from the model architecture. Cimpoi~\etal~\cite{FV-CNN} was one of the first works on the subject, where the authors compare the efficiency of two different CNN architectures for feature extraction: FC-CNN, which uses a fully connected (FC) layer, and FV-CNN, which uses a Fisher vector (FV)~\cite{FV} as a pooling method. They demonstrated that, in general, FC features are not that efficient because their output is highly correlated with the spatial order of the pixels. Later on, Condori and Bruno~\cite{condori2021} developed a model, called RankGP-3M-CNN, which performs multi-layer feature aggregation employing Global Average Pooling (GAP) to extract the feature vectors of activation maps at different depths of three combined CNNs (VGG-19, Inception-V3, and ResNet50). They propose a ranking technique to select the best activation maps given a training dataset, achieving promising results in some cases but at the cost of increased computational load, since three backbones are needed. Lyra~\etal~\cite{lyra2022} also proposes feature aggregation from multiple convolutional layers, but pooling is performed using an FV-based approach.

Numerous studies propose end-to-end architectures that enable fine-tuning of the backbone for texture recognition. Zhang~\etal~\cite{DeepTEN} proposed an orderless encoding layer on top of a DCNN, called Deep Texture Encoding Network (Deep-TEN), which allows images of arbitrary size. Xue~\etal~\cite{DEPNet} introduces a Deep Encoding Pooling Network (DEPNet), which combines features from the texture encoding layer from Deep-TEN and a global average pooling (GAP) to explore both the local appearance and global context of the images. These features are further processed by a bilinear pooling layer~\cite{lin2015}. In another work, Xue~\etal~\cite{xue2020} also combined features from differential images with the features of DEPNet into a new architecture. Using a different approach, Zhai~\etal~\cite{zhai2019} proposed the Multiple-Attribute-Perceived Network (MAP-Net), which incorporated visual texture attributes in a multi-branch architecture that aggregates features of different layers. Later on~\cite{zhai2020}, they explored the spatial dependency among texture primitives for capturing structural information of the images by using a model called Deep Structure-Revealed Network (DSRNet). Chen~\etal~\cite{chen2021} introduced the Cross-Layer Aggregation of a Statistical Self-similarity Network (CLASSNet). This CNN feature aggregation module uses a differential box-counting pooling layer that characterizes the statistical self-similarity of texture images. More recently, Yang~\etal~\cite{yang2022} proposed DFAEN (Double-order Knowledge Fusion and Attentional Encoding Network), which takes advantage of attention mechanisms to aggregate first- and second-order information for encoding texture features. Fine-tuning is employed in these methods to adapt the backbone to the new architecture along with the new classification head.

As an alternative to CNNs, Vision Transformers (ViTs)~\cite{Dosovitskiy2020} are emerging in the visual recognition literature. Some works have briefly explored their potential for texture analysis through the Describable Textures Dataset (DTD) achieving state-of-the-art results. Firstly, ViTs achieve competitive results compared to CNNs, but the lack of the typical convolutional inductive bias usually results in the need for more training data. To overcome this issue, a promising alternative is to use attention mechanisms to learn directly from text descriptions about images, \eg using Contrastive Language Image Pre-training (CLIP)~\cite{radford2021}. There have also been proposed bigger datasets for the pre-training of ViTs, such as Bamboo~\cite{zhang2022}, showing that these models scale well. Another approach is to optimize the construction of multitask large-scale ViTs such as proposed by Gesmundo~\cite{gesmundo2022continual} with the $\mu2$Net+ method.

%TODO
%\textbf{!!!Some important questions:}
% \begin{itemize}
%     \item \textbf{All of these methods are differentiable/fine-tuned?}
%     \item \textbf{All of them consider regular input sizes? (224x224)}
% \end{itemize}

\subsection{Randomized Neural Networks for Texture Analysis}

A Randomized Neural Network~\cite{schmidt1992feedforward,pao1992functional,pao1994learning,huang2006extreme}, in its simplest form, is a single-hidden-layer feed-forward neural network whose input weights are random, while the weights of the output layer are learned by a closed-form solution, in contrast to gradient-descent-based learning. Recently, several works have investigated RNNs to learn texture features for image analysis. Sá Junior~\etal~\cite{JarbasRNN2015} used small local regions of one image as inputs to an RNN, and the central pixel of the region as the target. The trained weights of the output layer for each image are then used as a texture representation. Ribas~\etal~\cite{ribas2018fusion} improved the previous approach with the incorporation of graph theory to model the texture image. Other works~\cite{junior2019randomized,ribas2022learning} have also extended these concepts to video texture analysis (dynamic texture).

The training of 1-layer RNNs as employed in previous works is a least-squares solution at the output layer. First, consider $X \in \mathbb{R}^{n \times z}$ as the input matrix with $n$ training samples and $z$ features, and $g=\phi(XW)$ as the forward pass of the hidden layer with a sigmoid nonlinearity, where $W \in \mathbb{R}^{z \times q}$ represents the random input weights for $q$ neurons. Given the desired output labels $Y$, the output weights $f$ are obtained as the least-squares solution
of a system of linear equations:
\begin{equation}\label{eq:lstsq}
f=Yg^{T}(gg^{T})^{-1}\,,     
\end{equation}
where $g^{T}(gg^{T})^{-1}$ is the Moore--Penrose 
pseudo-inverse~\cite{Moore1920,penrose_1955} of matrix~$g$.

An important aspect of RNNs is the generation of random weights for the first layer. Evidence suggests that this choice has little impact once the weights are fixed. In this sense, a common trend among previous works is the use of the Linear Congruential Generator (LCG), a simple pseudo-random number generator in the form of $x_{k+1} = (ax_{k} + b) \mod c$.

%\textbf{WE SHOULD SKIP THE DESCRIPTION OF RNNS, AND EXPLAIN ONLY AUTO-ENCODERS}

% In the RNN, the input features are mapped to a $Q-$dimensional random feature space (number of hidden neurons),

% \begin{equation}
%     Z=\phi(WX)
% \end{equation}
% where $\phi(.)$ is the hidden node activation function, $W$ is the random weight matrix between the input and hidden layers whose dimensions are  $Q\times(p+1)$, and $X$ is the input feature matrix $X=\left[\vec{x_1}, \vec{x_2},\ldots, \vec{x_N}\right]$, composed of $N$ input vectors with $p$ attributes. Given the $Z$ matrix of random hidden features (hidden node outputs), the RNN calculates the output weights $\beta$ from

% \begin{equation}
%     \beta=YZ^{T}(ZZ^{T})^{-1},
% \end{equation}

% where $Z^{T}(ZZ^{T})^{-1}$ is the Moore-Penrose pseudo-inverse of matrix $Z$ \cite{Moore1920,penrose_1955}
% and $Y=\left[\vec{y_i}, \vec{y_2},\ldots, \vec{y_N}\right]$ is a matrix of labels (each $\vec{y_i}$ corresponding to its respective input vector $\vec{x_i}$). 

The RNN can be used as a randomized autoencoder (RAE)~\cite{RNN-AE} by considering the input feature matrix $X$ as the target output $Y=X$. In this sense, the model is composed of a random encoder and a least-squares-based decoder that can map the input data. Kasun~\etal~\cite{RNN-AE} also suggests the use of random orthogonal weights~\cite{saxe2013exact} for the initialization of the encoder. In this way, the weight matrix $f$ will represent the transformation of the projected random space back into the input data $X$ (output).

% Randomized neural networks \citep{schmidt1992feedforward,pao1992functional,pao1994learning,huang2006extreme} are artificial neural nets, which, in their simplest version, have a single hidden layer, whose weights are determined randomly, and an output layer whose weights can be computed using a closed-form solution. When these neural networks allow direct links between the input feature vectors and the output layer \citep{pao1992functional,pao1994learning}, they are known as random vector functional link (RVFL) nets.

\section{RADAM for Texture Feature Encoding}
%The majority of the previous works on deep texture recognition consider end-to-end models that couple new modules around a pre-trained backbone, and involves fine-tuning the new architecture on specific texture recognition tasks. Here, we propose a new feature encoding module that acts on the backbone only as a feature extractor, and a dedicated classifier is applied over the obtained features (no backbone fine-tuning is done). The method, called RADAM, is described in the following.
%The main concepts of the proposed RADAM method consist of multi-depth feature aggregation and randomized pixel-wise encoding to compose a single feature vector, given an input image processed by the backbone. 

%The module itself is not trained via conventional gradient descent, and the backbone is frozen: no fine-tuning or backpropagation is performed (only a linear classifier is trained at the top). We named the method  \textbf{Ra}ndom Encoding of \textbf{D}eep \textbf{A}ctivation \textbf{M}aps (RADAM), which we describe in the following.

The main idea of the proposed RADAM method is to use multi-depth feature aggregation and randomized pixel-wise encoding to compose a single feature vector, given an input image processed by the backbone. First of all, consider an input image $I \in \mathbb{R}^{w_0 \times h_0 \times 3}$ fed into a backbone $B= (d_1,...,d_n)$, consisting of $n$ blocks of convolutional layers. 
% shouldn't it rather be $B= ( ..)$, as a set has no ordering
An activation map, i.e., the output of any convolutional block given $I$, is a 3-dimensional tensor (ignoring the batch dimension, for simplicity) $X_i \in \mathbb{R}^{w_i \times h_i \times z_i}$. The process of feature aggregation consists of combining the outputs of different activation maps at different depths. To that end, we divide the backbone into a fixed number of blocks according to different depths. This division is made to keep a fixed number of blocks for feature extraction, regardless of the total depth of the backbone architecture.

\subsection{Pre-trained Deep Convolutional Networks: Backbone selection}

Most previous works on texture analysis consider pre-trained ResNets~\cite{he2016deep} (18 or 50) as backbones. Here, we consider the output of five blocks of layers according to the ResNet architecture, meaning that five activation maps are considered for feature aggregation. Additionally, we consider the ConvNeXt architecture~\cite{liu2022ConvNeXt}, a more recent method with promising results in image recognition. For this backbone, we consider the activation maps from the four blocks of layers according to the architecture described in the original work. More specifically, the following ConvNeXt configurations are used, with their corresponding number of channels ($z_i$) of each block:
\begin{itemize}
    \item ConvNeXt-nano \footnote{This variant was not presented in the original paper, but is available at~\url{https://github.com/rwightman/pytorch-image-models/blob/main/timm/models/ConvNeXt.py}}: $z_i= (80, 160, 320, 640)$.
    \item ConvNeXt-T: $z_i= (96, 192, 384, 768)$. 
    \item ConvNeXt-B: $z_i= (128, 256, 512, 1024)$
    \item ConvNeXt-L: $z_i= (192, 384, 768, 1536)$.
    \item ConvNeXt-XL: $z_i= (256, 512, 1024, 2048)$.
\end{itemize}

As a lightweight alternative, MobileNet V2~\cite{sandler2018mobilenetv2}
 is also compared within our experiments (with five blocks of layers) using either the original architecture or a 1.4 width multiplier, i.e., $z_i= (16, 24, 32, 96, 320)$ and $z_i= (24, 32, 48, 136, 448)$, respectively.

\subsection{Deep Activation Map Preparation}

Given each deep activation map $X_i$, we apply a depth-wise $l_p$-normalization ($p=2$, i.e., Euclidean norm)
\begin{equation}\label{eq:norm}
    X_i(:,:,j) = \frac{X_i(:,:,j)}{\max(||X_i(:,:,j)||_2)}, 
\end{equation}
 where $X_i(:,:,j)$ represents the 2-dimensional activation map at each channel $j \in z_i$ with spatial sizes $(w_i, h_i)$.
 
\begin{figure}[!htb]
  \centering
  \subfigure[The proposed feature encoding module (RADAM).]{ 
    \includegraphics[width=0.65\linewidth]{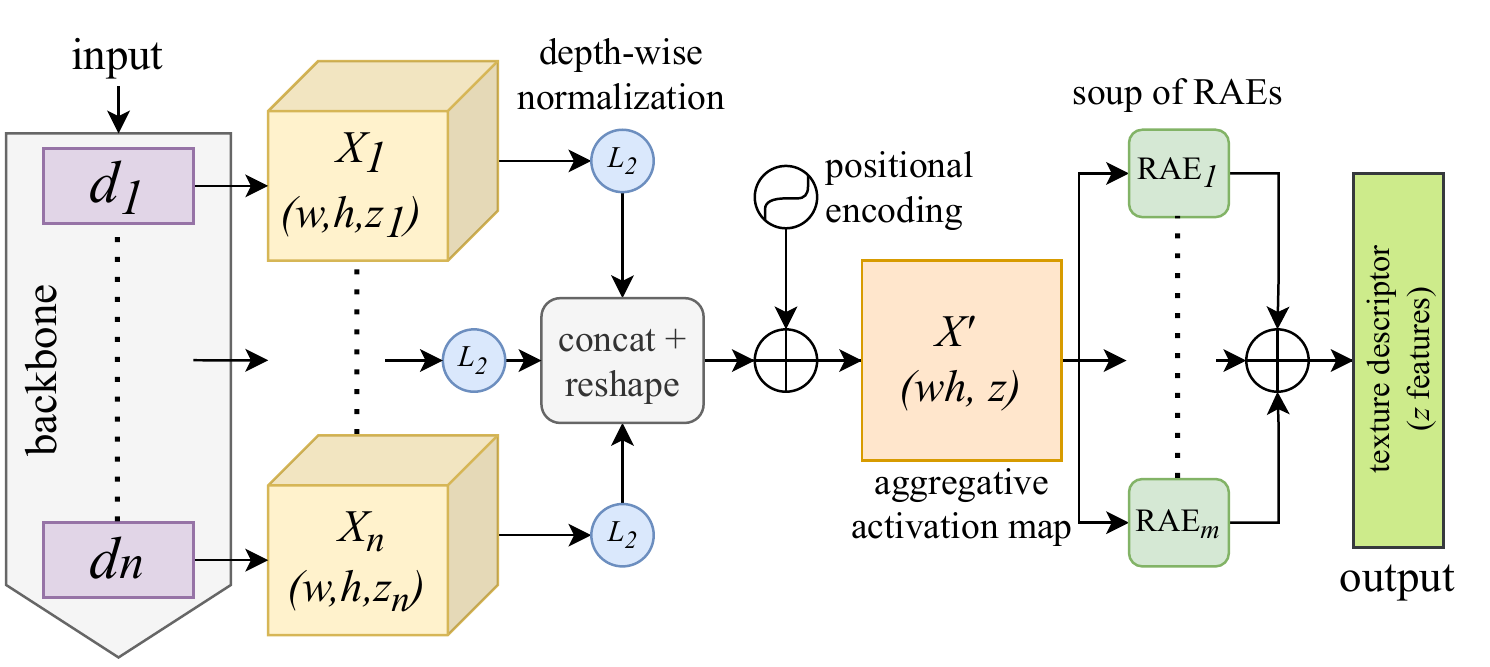}
    \label{fig:method-a}
     }
  \subfigure[Randomized Auto-encoder (RAE).]{
    \includegraphics[width=0.275\linewidth]{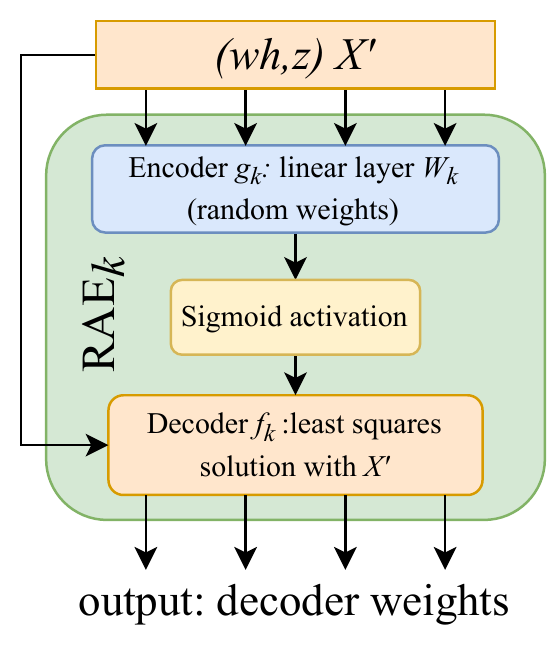}
    \label{fig:method-b}
  }
  \caption{Illustration of the proposed RADAM architecture (a), given an input image to a final descriptor. The RAE is shown in detail (b), which is a simple 1-layer auto-encoder solved through least-squares, where we use the decoder weights as descriptors (summed for $m$ RAEs).}
  \label{fig:method}
\end{figure}

For feature aggregation, we propose to concatenate the activation maps along the third dimension ($z_i$). However, each map $X_i$ initially has a different spatial dimension $w_i$ and $h_i$. To overcome this, we simply resize all activation maps with bilinear interpolation using the spatial dimensions of $X_{\frac{n}{2}}$, $(w_{\frac{n}{2}}, h_\frac{n}{2})$, as the target sizes. In other words, we consider the spatial dimensions at the middle of the backbone as our anchor size, meaning that some activation maps will require upscaling (if $i > \frac{n}{2}$) and others downscaling (if $i < \frac{n}{2}$). Naturally, the information from activation maps at higher depths receives higher priority considering that upscaling preserves more information than downscaling. These assumptions consider the most common structure of convolutional architectures where the spatial size decreases with layer depth. Nonetheless, the idea is to keep all activation maps with a fixed spatial dimension. From now on, we will refer to spatial dimensions of all $X_i$ as $w=w_{\frac{n}{2}}$ and $h=h_{\frac{n}{2}}$. For an input size of 224x224, this results in $w=h=28$ for the backbones explored in this work. The concatenation of activation maps is then performed as
\begin{equation}
    X' = [X_1 ; ... ; X_n ] \in \mathbb{R}^{w \times h \times z} \to X'  \in \mathbb{R}^{wh \times z}\,,
\end{equation}
 where $[.;.]$ denotes the concatenation along the third dimension, and $z=\sum_i z_i$ is the resulting number of channels after concatenation. Considering common convolutional architectures where $z_{i} < z_{i+1}$, activation maps from higher depths have a higher influence on the overall $z$ features. Additionally, the 2-dimensional activation map at each channel $z_i$ is flattened, resulting in the reshaped 2D representation $X'$ with sizes $wh$-by-$z$, which we refer to as an aggregated activation map. These steps are illustrated in~\cref{fig:method-a}, which reports the overall structure of the proposed method.

% RAE in the different parts of the figure should be in roman, not italcs
% in the last RAE_k, k should be in italics

\subsection{Pixel-wise Randomized Encoding}

The aggregated activation map of a single image is used to train an RAE considering each spatial point, or pixel (row of $X'$), as a sample and each channel (column of $X'$) as a feature. 
In this sense, the method also works with arbitrary input sizes (if accepted by the backbone) since the spatial dimensions only affect the number of training samples for the RAE. Intuitively, larger input sizes would improve the RAE training, but would also increase the backbone cost significantly. Therefore, in this work, we consider only a constant input size of 224x224 (forced resizing), since this is the most common configuration of various backbones. Moreover, considering that the spatial organization of the pixels is lost due to the flattening procedure of $X'$, we add a positional encoding composed of sine and cosine functions of different frequencies with dimension $z$ as proposed in~\cite{vaswani2017attention}, extended for 2 spatial dimensions as in~\cite{wang2021translating}:
\begin{align*}
    &\PE(x,y, 2i) = \sin (\frac{x}{10000^{4i/z}}), \\ 
    &\PE(x,y, 2i + 1) = \cos (\frac{x}{10000^{4i/z}}), \\
    &\PE(x,y, 2j + z/2) = \sin (\frac{y}{10000^{4j/z}}), \\
    &\PE(x,y, 2j + 1 + z/2) = \cos (\frac{y}{10000^{4j/z}}), \\
\end{align*}
where $x \in w$ and $y \in h$, which is then added to the aggregated activation map via element-wise sum:
\begin{equation}
    X' = X' \oplus PE.
\end{equation}
% D has not ben explained, I also don't see how the last equation connects to the preceding, is this an equation or a replacement of X' (more like an instruction in your code
%my mistake, D is z in our case. The last equation only connects to PE in the sense that we sum it, I included an additional sentence there

After summing the positional encoding to $X'$, the first step of the RAE is to project the inputs using a random fully-connected layer with weights $W_k \in  \mathbb{R}^{z \times q}$, followed by a sigmoid nonlinearity. The weights are generated using the LCG for simplicity and better replicability, followed by standardization (zero centered, unity variance) and orthogonalization~\cite{saxe2013exact}. These configurations were chosen according to previous works~\cite{ribas2018fusion,RNN-AE}. As for the LCG parameters, we use $a=75$, $b=74$, and $c=2^{16}+1$, starting with $x=0$, which is a classical configuration according to the ZX81 computer from 1981. Here $k$ works like a seed for random sampling, denoting a starting index inside the LCG space generated with the given configuration. More details on LCG weights are given in the Supplementary Material, such as an ablation on the impacts of different LCG configurations. The forward pass of the encoder $g_k \in \mathbb{R}^{wh \times q}$ for all samples is then obtained as
\begin{equation}
    g_k = \phi(X'W_k)\,,  
\end{equation}
and the decoder weights $f_k \in \mathbb{R}^{z \times q}$ are obtained as the least-squares solution described in~\cref{eq:lstsq}, changing the target $Y$ to $X'$:
\begin{equation}
    f_k = X'g_k^{T}(g_kg_k^{T})^{-1}\,.
\end{equation}

The main idea of employing an individual randomized neural network for each image is to use the output weights themselves as a representation. In the case of RAEs, the output layer has the same dimension as the input layer. Therefore, a single hidden neuron ($q=1$) is considered to maintain the dimensionality. In this sense, the resulting decoder weights are represented by
\begin{equation}
    f_k = (\nu_1, \ldots, \nu_z)\,, 
\end{equation}
where $\nu_i$ represents the connection weight between the single hidden neuron and the output $i$, corresponding to feature $i \in z$.

A single-neuron RAE may be limited in encoding enough information contained in the deep activation maps. Therefore, we propose an ensemble of models or, as recently introduced~\cite{wortsman2022model}, a model ``soup'', which is achieved by combining the weights of $m$ parallel models. Here, each model is an RAE with a different random encoder (using a different LCG seed), and the combination is performed by summing the decoder weights
\begin{equation}
    \varphi_m = (\sum_{k=1}^m f_k(\nu_1) ,\ldots, \sum_{k=1}^m f_k(\nu_z))\,.
\end{equation}

It is important to note that the encoders $g_k$ of each of the $m$ RAEs have a different random weight initialization. This is achieved by creating an LCG sequence of size $mz$ so that we have $z$ weights for each of the $m$ RAEs. The structure of the RAE is illustrated in~\cref{fig:method-b}, and following the whole RADAM pipeline shown in~\cref{fig:method-a}, a texture representation, or feature vector $\varphi_m$, is obtained for the input image. The code for all these steps is available in the Supplementary Material and in our online repository~\textsuperscript{\ref{github}}. The feature vectors $\varphi_m$ are then used to train a linear classifier for a given texture recognition task (more details on the classifier can be consulted in \cref{sec:classifiers}). 

\section{Experiments and Results}\label{sec:experiments}

% \subsection{Setup}\label{sec:setup_bkb}
\subsection{Setup}

Our model is implemented using PyTorch~\cite{Paszke_PyTorch_An_Imperative_2019} (except for the classification step), making it easier to couple RADAM with several methods implemented in this library. The classification step is performed using Scikit-learn~\cite{scikit-learn}.  
%with a linear SVM with $C=1$, a maximum of 100k iterations, and keeping all other standard parameters.
We measure our results by the average classification accuracy and corresponding standard deviation, when applicable (depending on the dataset). For the backbones, we consider the PyTorch Image Models library~\cite{rw2019timm} (version 0.6.7), which contains several pre-trained computer vision methods. In the Supplementary Material, we present the main code for RADAM, and the complete implementation including scripts for experimentation can be consulted in our GitHub repository~\textsuperscript{\ref{github}}.

Seven texture datasets are used for evaluation purposes in this paper, of which the following two variants of the Outex dataset~\cite{outex} were used for analyzing the RADAM method alone: \begin{itemize}
    \item \textit{Outex10}: Composed of 4320 grayscale images in 24 different texture classes. This dataset focuses on rotation invariance;
    \item \textit{Outex13}: This suite holds 1360 RGB images in 68 texture classes, and evaluates color texture recognition.
    %\item \textit{Outex14}: It focuses on RGB images with illumination invariance and it contains 2040 images divided into 68 different classes.
\end{itemize}

The following five datasets are used for comparisons with other methods:
\begin{itemize}
    \item \textit{Describable Texture Dataset (DTD)}~\cite{FV}: Composed of 5640 images in 47 different texture classes, evaluated by the 10 provided splits for training, validation, and test;
    \item \textit{Flickr Material Dataset (FMD)}~\cite{sharan2010}: Holds 1000 images representing 10 material categories, and validation is done through 10 repetitions of 10-fold cross-validation; 
    \item \textit{KTH-TIPS2-b}~\cite{caputo2005}: Contains 4752 images of 11 different materials. This dataset has a fixed set of 4 splits for 4-fold cross-validation;
    \item \textit{Ground Terrain in Outdoor Scenes (GTOS)}~\cite{xue2020}: This dataset represents 34105 images divided into 40 outdoor ground materials classes. There is also a fixed set of 5 train/test splits;
    \item \textit{GTOS-Mobile} \cite{DEPNet}: Consists of 100011 images captured from a mobile phone of 31 different outdoor ground materials, and contains a single train/test split.
\end{itemize} 

\subsection{Analysis of RADAM properties}

Our first experimental evaluation concerns aspects of the proposed RADAM method. In the Supplementary Material, we show an additional analysis of the impacts caused by different random weights (LCG configurations) and concluded that they are minimal, corroborating previous works. In the following, we evaluate and discuss other aspects of RADAM.

\subsubsection{Positional encoding and different classifiers}\label{sec:classifiers}

We evaluate two design choices for the RADAM pipeline, the use of positional encoding and the classifier. The method is compared with or without encoding under two different classifiers: Linear Discriminant Analysis (LDA)~\cite{ripley2007pattern} and Support Vector Machines (SVM)~\cite{platt1999probabilistic}. For LDA the least-squares solution with automatic shrinkage using the Ledoit--Wolf lemma is used, and a linear kernel with $C=1$ is considered for SVM. Since the evaluation of positional encoding concerns the spatial properties of texture, we consider the Outex10 benchmark, which focuses on rotation invariance. As the results in~\cref{tab:encoding} demonstrate, positional encoding improves or maintains performance in all cases, especially under the LDA classifier. On the other hand, SVM provides the best results in all cases while gaining less improvement from positional encoding. Nevertheless, we keep the positional encoding in our architecture since the additional cost is negligible compared to the potential gains. The SVM is also used as the classifier for all the following experiments.

\begin{table}[!htb]
	\centering
	\caption{\label{tab:encoding}Ablations with positional encoding and different classifiers, using RADAM ($m=1$) with different backbones. Results are measured by classification accuracy considering the single train/test split of Outex10.}
	
	\begin{tabular}{cc|cc}
		\hline
		
    	 &&  \multicolumn{2}{c}{Outex10}  \\
    	 encoding&backbone & LDA & SVM\\
    	\hline
\hline
none&ResNet18&77.5&83.4\\ 
positional&ResNet18&79.5&84.7\\ 

 \hline 
none&ConvNeXt-nano&82.4&89.6\\ 
positional&ConvNeXt-nano&85.9&89.6\\ 

 \hline 
none&ResNet50&86.0&87.4\\ 
positional&ResNet50&87.4&87.4\\ 

 \hline 
none&ConvNeXt-T&87.8&91.1\\ 
positional&ConvNeXt-T&89.4&91.1\\

    	\hline
	\end{tabular}
\end{table}

\subsubsection{Soup size}

\begin{figure}[!htb]
  \centering
  
    \includegraphics[width=0.5\linewidth]{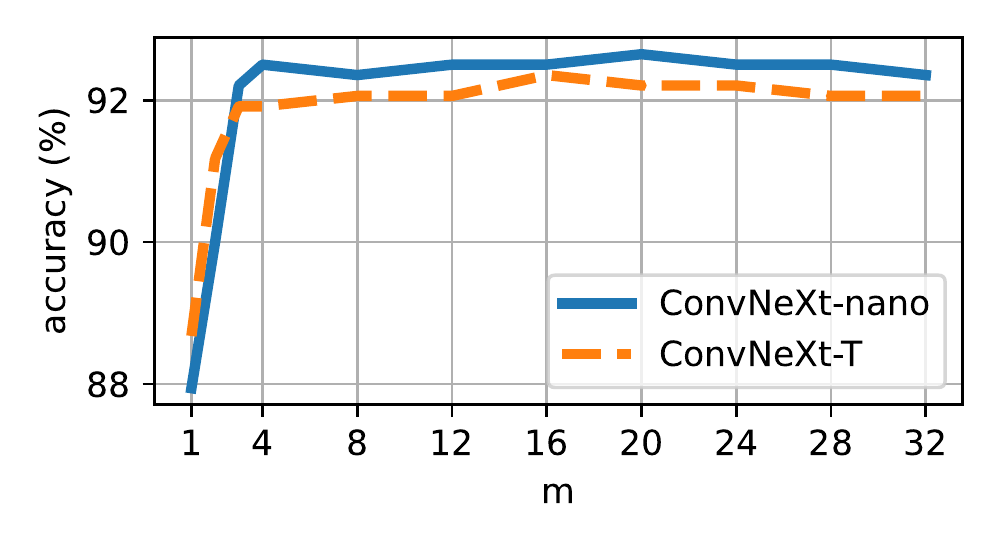}

%   \hfill
%   \begin{subfigure}{0.7\linewidth}
%     \includegraphics[width=1.0\linewidth]{Mablation-norms-L2spatial_RAEspatial_Outex14.pdf}
%     \caption{Outex14.}
%     \label{fig:m-b}
%   \end{subfigure}
  \caption{Impacts of changing the number of RAEs ($m$) in the proposed method, considering different backbones on the Outex13 dataset. All other experiments in this paper consider $m=4$.}
  \label{fig:m}
\end{figure}

The only free parameter of the proposed RADAM method is the number of RAEs to be combined, i.e., $m$. We evaluated $m$ ranging from 1 to 32 and the results are shown in \cref{fig:m} for different backbones in the Outex13 dataset. We observe significant gains for $m$ from 1 to 4, while for larger values performance tends to stabilize. These results indicate that $4 \leq m \leq8$ is a good approach for a balance between performance and cost since no significant gains are achieved above that. All the following experiments in this paper are performed using $m=4$.

The effects observed when increasing $m$ are expected considering what is usually seen in model ensembles, or ``model soups''~\cite{wortsman2022model}, where the combination of models trained separately may be beneficial. On the other hand, our encoders are random, and each one has different weights. However, even if each encoder creates a different random projection of the input, the decoders learn to transform the projection back to the same feature space. In other words, the RAEs learn different encoding-decoding functions for the same input that, when combined, provide a better representation in our feature extraction use case.

\subsection{Comparison with other pooling techniques}

To show the gains of RADAM over the common pooling approach, we performed additional experiments using Global Average Pooling (GAP) coupled with SVM using the same configurations considered for RADAM. We use GAP applied over the output of the last layer of the backbone (the usual approach in most CNN architectures), and also GAP agg., which aggregates the GAP from each of the $n$ feature blocks and returns an image representation with $z$ features (as RADAM). The results are shown in~\cref{tab:poolings}, where it is possible to observe that RADAM overcomes these two approaches by a considerable margin, in all backbones and datasets. GAP agg. usually improves over the regular GAP of only the last convolutional layer, which is to be expected since more features are being added. However, the regular GAP sometimes performs better than this simple aggregation by concatenation. On the other hand, the gains when using RADAM for aggregation are far more expressive and result in state-of-the-art performance in all benchmarks we considered, as we discuss in the following section.

\begin{table}[!htb]
	\centering
	\caption{\label{tab:poolings}Classification accuracy (linear SVM) when using image features obtained from the backbones with different techniques: the GAP, the aggregation of GAP from each of the $n$ feature blocks of the backbones (the same way we approach them with RADAM), and the proposed RADAM. We consider two challenging texture benchmarks (DTD and FMD).}
	% \resizebox{\linewidth}{!}{
	\begin{tabular}{cc|cc}
		\hline
		
    	method & backbone & DTD & FMD \\
     
        % method & backbone & Outex10 & Outex13\\
        \hline \hline
        % SOTA &ResNet50& \textbf{77.6}{\tiny$\pm$0.6} & \textbf{86.9}\\

GAP &ResNet18&64.1{\tiny$\pm$1.1}&77.1{\tiny$\pm$0.7}\\ 
GAP agg. &ResNet18&64.6{\tiny$\pm$1.2}&77.2{\tiny$\pm$0.6}\\ 
RADAM&ResNet18&68.1{\tiny$\pm$1.0}&77.7{\tiny$\pm$0.5}\\ 

 \hline 
GAP &ConvNeXt-nano&61.9{\tiny$\pm$1.0}&73.8{\tiny$\pm$0.9}\\ 
GAP agg. &ConvNeXt-nano&71.7{\tiny$\pm$0.6}&84.4{\tiny$\pm$0.4}\\ 
RADAM&ConvNeXt-nano&74.9{\tiny$\pm$0.7}&87.1{\tiny$\pm$0.4}\\ 

 \hline 
GAP &ResNet50&72.6{\tiny$\pm$0.9}&84.5{\tiny$\pm$0.6}\\ 
GAP agg. &ResNet50&74.6{\tiny$\pm$0.6}&83.4{\tiny$\pm$0.4}\\ 
RADAM&ResNet50&75.6{\tiny$\pm$1.1}&85.3{\tiny$\pm$0.4}\\ 

 \hline 
GAP &ConvNeXt-T&66.0{\tiny$\pm$1.0}&79.0{\tiny$\pm$0.4}\\ 
GAP agg. &ConvNeXt-T&73.7{\tiny$\pm$0.9}&83.6{\tiny$\pm$0.5}\\ 
RADAM&ConvNeXt-T&77.0{\tiny$\pm$0.7}&88.7{\tiny$\pm$0.4}\\ 

 \hline 
GAP &ConvNeXt-T in ImageNet-21K&78.4{\tiny$\pm$0.7}&91.4{\tiny$\pm$0.3}\\ 
GAP agg. &ConvNeXt-T in ImageNet-21K&77.3{\tiny$\pm$0.9}&89.9{\tiny$\pm$0.5}\\ 
RADAM&ConvNeXt-T in ImageNet-21K&81.4{\tiny$\pm$0.7}&93.0{\tiny$\pm$0.3}\\ 

 \hline 
GAP &ConvNeXt-B&69.2{\tiny$\pm$1.3}&83.6{\tiny$\pm$0.5}\\ 
GAP agg. &ConvNeXt-B&73.5{\tiny$\pm$1.0}&86.7{\tiny$\pm$0.5}\\ 
RADAM&ConvNeXt-B&76.4{\tiny$\pm$0.9}&90.2{\tiny$\pm$0.2}\\ 

 \hline 
GAP &ConvNeXt-B in ImageNet-21K&79.2{\tiny$\pm$0.6}&91.9{\tiny$\pm$0.4}\\ 
GAP agg. &ConvNeXt-B in ImageNet-21K&80.4{\tiny$\pm$1.0}&92.3{\tiny$\pm$0.3}\\ 
RADAM&ConvNeXt-B in ImageNet-21K&82.8{\tiny$\pm$0.9}&94.0{\tiny$\pm$0.2}\\ 

 \hline 
GAP &ConvNeXt-L&70.9{\tiny$\pm$0.9}&84.6{\tiny$\pm$0.4}\\ 
GAP agg. &ConvNeXt-L&73.4{\tiny$\pm$0.6}&86.4{\tiny$\pm$0.5}\\ 
RADAM&ConvNeXt-L&77.4{\tiny$\pm$1.1}&89.3{\tiny$\pm$0.3}\\ 

 \hline 
GAP &ConvNeXt-L in ImageNet-21K&80.4{\tiny$\pm$0.9}&92.2{\tiny$\pm$0.4}\\ 
GAP agg. &ConvNeXt-L in ImageNet-21K&81.9{\tiny$\pm$0.6}&93.4{\tiny$\pm$0.4}\\ 
RADAM&ConvNeXt-L in ImageNet-21K&84.0{\tiny$\pm$1.0}&95.2{\tiny$\pm$0.4}\\ 

 \hline 
GAP &ConvNeXt-XL in ImageNet-21K&81.3{\tiny$\pm$1.0}&92.4{\tiny$\pm$0.3}\\ 
GAP agg. &ConvNeXt-XL in ImageNet-21K&82.0{\tiny$\pm$1.1}&93.9{\tiny$\pm$0.4}\\ 
RADAM&ConvNeXt-XL in ImageNet-21K&83.7{\tiny$\pm$0.9}&95.2{\tiny$\pm$0.3}\\ 

 \hline 

	\end{tabular}
	% }
\end{table}

\subsection{Comparison with literature}

Finally, we compare RADAM with several state-of-the-art methods on five challenging texture recognition datasets; all results are shown in \cref{tab:final-results}. The table is organized into separate rows according to the different backbones in terms of computational budget. We indicate the pre-training dataset used for the backbone; ImageNet-1K~\cite{deng2009imagenet} was used when not stated. Firstly, we show the results obtained using RADAM and MobileNet V2 (standard or with width multipliers of 1.4) as a lightweight alternative (costs are discussed in depth in~\cref{sec:costs}). The first comparison section contains methods using ResNet18, and we also included RADAM with ConvNeXt-nano in this section considering that it has a similar computational budget. In this scenario, RADAM achieves competitive performance on KTH using ResNet18 but is less effective on other datasets. On the other hand, RADAM with MobileNetV2 1.4 achieves better results than ResNet18 in all cases except GTOS-Mobile and beats all the ResNet18 literature methods on DTD, FMD, and KTH, proving to be an excellent low-cost approach. Using ConvNeXt-nano, RADAM also achieves the best results on all datasets except GTOS and GTOS-Mobile.

\begin{table*}
	\centering
	\caption{\label{tab:final-results}Classification accuracy of different methods on texture benchmarks. The used backbones are separated into row blocks according to their computational budget, the input size is indicated in parentheses (224x224 when not stated), and the two best results in each block are highlighted in bold type. Results in \textcolor{blue}{blue} show the previous state-of-the-art on each dataset, and \textcolor{red}{red} represents our results matching or above that.}
	\resizebox{\linewidth}{!}{
	\begin{tabular}{cc|ccccc}
		\hline
		
    	method & backbone & DTD & FMD & KTH-2-b & GTOS & GTOS-Mobile\\
        \hline \hline

RADAM light (ours)&MobileNet V2&69.3{\tiny$\pm$0.8}&80.5{\tiny$\pm$0.6}&84.8{\tiny$\pm$0.5}&81.0{\tiny$\pm$1.3}&75.1\\ 

RADAM light (ours)&MobileNet V2 1.4&73.1{\tiny$\pm$0.9}&82.6{\tiny$\pm$0.5}&86.8{\tiny$\pm$3.1}&81.7{\tiny$\pm$1.7}&78.2\\

 \hline

        DeepTEN & ResNet18 (352) & & & & &76.1 \\
        DEPNet  & ResNet18& & & & &82.2 \\
        MAPNet& ResNet18 & 69.5{\tiny$\pm$0.8} & 80.8{\tiny$\pm$1.0} & 80.9{\tiny$\pm$1.8} &80.3{\tiny$\pm$2.6}& 83.0 \\
    	DSRNet&ResNet18& 71.2{\tiny$\pm$0.7} &81.3{\tiny$\pm$0.8}&81.8{\tiny$\pm$1.6}&81.0{\tiny$\pm$2.1}& \textbf{83.7 } \\
        CLASSNet&ResNet18& \textbf{71.5}{\tiny$\pm$0.4} &\textbf{82.5}{\tiny$\pm$0.7}&\textbf{85.4}{\tiny$\pm$1.1}&\textbf{84.3}{\tiny$\pm$2.2}& \textbf{85.3}\\
        
        RADAM (ours)&ResNet18&68.1{\tiny$\pm$1.0}&77.7{\tiny$\pm$0.5}&84.7{\tiny$\pm$3.6}&80.6{\tiny$\pm$1.7}&79.5\\

        RADAM (ours)&ConvNeXt-nano&\textbf{74.9}{\tiny$\pm$0.7}&\textbf{87.1}{\tiny$\pm$0.4}&\textbf{89.6}{\tiny$\pm$3.8}&\textbf{83.7}{\tiny$\pm$1.5}&81.8\\

      \hline 

        DeepTEN&ResNet50 (352) & 69.6  &80.2{\tiny$\pm$0.9}&82.0{\tiny$\pm$3.3}&84.5{\tiny$\pm$2.9}&\\
        MAPNet& ResNet50& 76.1{\tiny$\pm$0.6} & 85.2{\tiny$\pm$0.7} & 84.5{\tiny$\pm$1.3} & 84.7{\tiny$\pm$2.2}& 86.6 \\
        DSRNet&ResNet50& \textbf{77.6}{\tiny$\pm$0.6} &86.0{\tiny$\pm$0.8}&85.9{\tiny$\pm$1.3}&85.3{\tiny$\pm$2.0}& \textcolor{blue}{\textbf{87.0}} \\
        CLASSNet&ResNet50& 74.0{\tiny$\pm$0.5} &86.2{\tiny$\pm$0.9}&87.7{\tiny$\pm$1.3}& \textcolor{blue}{\textbf{85.6}}{\tiny$\pm$2.2}& 85.7 \\
        DFAEN&ResNet50& 73.2 &86.9& 86.3 & &\textbf{86.9} \\
        RADAM (ours)&ResNet50&75.6{\tiny$\pm$1.1}&85.3{\tiny$\pm$0.4}&88.5{\tiny$\pm$3.2}&81.8{\tiny$\pm$1.1}&81.0\\ 
        
        RADAM (ours)&ConvNeXt-T&77.0{\tiny$\pm$0.7}&\textcolor{red}{\textbf{88.7}}{\tiny$\pm$0.4}&\textbf{90.7}{\tiny$\pm$4.0}&84.2{\tiny$\pm$1.7}&85.3\\
        
         RADAM (ours)&ConvNeXt-T in ImageNet-21K&\textbf{81.4}{\tiny$\pm$0.7}&\textcolor{red}{\textbf{93.0}}{\tiny$\pm$0.3}&\textbf{91.0}{\tiny$\pm$4.9}&\textbf{85.4}{\tiny$\pm$1.6}&86.5\\ 

 \hline 

DFAEN&Densenet161&76.1 &87.6& 86.6 & &86.9\\
Multilayer-FV&EfficientNet-B5 (512)&78.9& \textcolor{blue}{88.7} &82.9 &&  \\

        RankGP-3M-CNN++ & (3 backbones) & & 86.2{\tiny$\pm$1.4} & \textcolor{blue}{91.1}{\tiny$\pm$4.5} && \\
         fine-tuning& ViT B/16 in Bamboo 69m &81.2 &&&&\\ 
        CLIP zero-shot & ViT L/14 (336) in WIT 400m & \textcolor{blue}{83.0} &&&&\\
        
       $\mu$2Net+ & ViT-L/16 (384) in ImageNet-21K& 82.2 &&&&\\

        RADAM (ours)&ConvNeXt-B&76.4{\tiny$\pm$0.9}&\textcolor{red}{90.2}{\tiny$\pm$0.2}&87.7{\tiny$\pm$5.6}&84.1{\tiny$\pm$1.6}&82.2\\
        
        RADAM (ours)&ConvNeXt-B in ImageNet-21K&82.8{\tiny$\pm$0.9}&\textcolor{red}{94.0}{\tiny$\pm$0.2}&\textcolor{red}{\textbf{91.8}}{\tiny$\pm$4.1}&\textcolor{red}{\textbf{86.6}}{\tiny$\pm$1.7}&\textcolor{red}{87.1}\\

        RADAM (ours)&ConvNeXt-L&77.4{\tiny$\pm$1.1}&\textcolor{red}{89.3}{\tiny$\pm$0.3}&89.3{\tiny$\pm$3.4}&84.0{\tiny$\pm$1.8}&85.8\\ 
        
        RADAM (ours)&ConvNeXt-L in ImageNet-21K&\textcolor{red}{\textbf{84.0}}{\tiny$\pm$1.0}&\textcolor{red}{\textbf{95.2}}{\tiny$\pm$0.4}&\textcolor{red}{91.3}{\tiny$\pm$4.1}&\textcolor{red}{85.9}{\tiny$\pm$1.6}&\textcolor{red}{\textbf{87.3}}\\
        
        RADAM (ours)&ConvNeXt-XL in ImageNet-21K&\textcolor{red}{\textbf{83.7}}{\tiny$\pm$0.9}&\textcolor{red}{\textbf{95.2}}{\tiny$\pm$0.3}&\textcolor{red}{\textbf{94.4}}{\tiny$\pm$3.8}&\textcolor{red}{\textbf{87.2}}{\tiny$\pm$1.9}&\textcolor{red}{\textbf{90.2}}\\

		\hline
	\end{tabular}
	}
\end{table*}

Considering the results within the ResNet50 budget, RADAM performs much better compared to ResNet18. Competitive results are achieved on most datasets using ResNet50, and also the best results on KTH. Using ConvNeXt-T RADAM achieves state-of-the-art on FMD, and also overcomes the compared methods on KTH. Performance is improved even further considering ConvNeXt-T pre-trained on ImageNet-21K, achieving the best results on most of the datasets. These results show the potential gains of better pre-training of the employed backbone.

The results shown in the last block of rows in~\cref{tab:final-results} concern methods with an increased cost compared to the previous ones. Here we compare RADAM using ConvNeXt-B, L, and XL against several works including very recent methods, such as ViTs. Our method achieves state-of-the-art results on all datasets in this case, especially with the ConvNeXt-L/XL backbones. It is possible to notice again that the better pre-training with ImageNet-21K results in significant performance increases.

\subsection{Feature extraction cost versus performance}\label{sec:costs}

\begin{figure}[!htb]
  \subfigure[Best methods on the DTD dataset. Text inside the plot indicates backbones used with RADAM.]{
  \centering
    \includegraphics[width=0.33\linewidth]{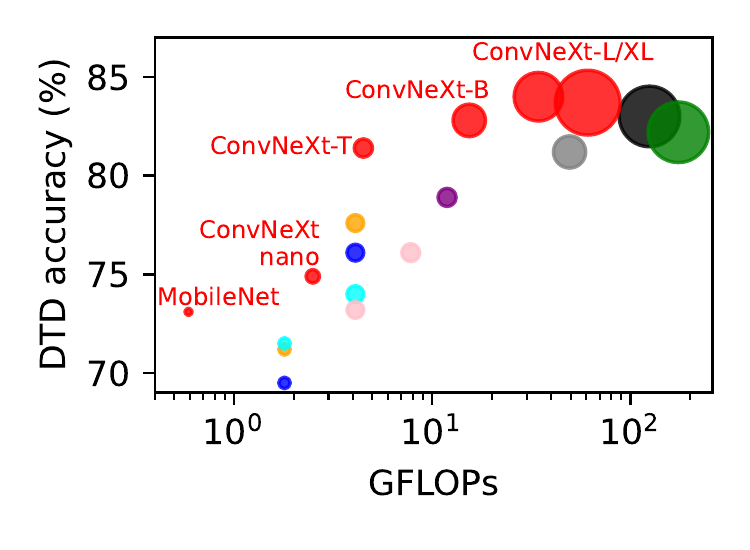}
    \label{fig:cost-dtd}
  }
  \includegraphics[width=0.3\linewidth]{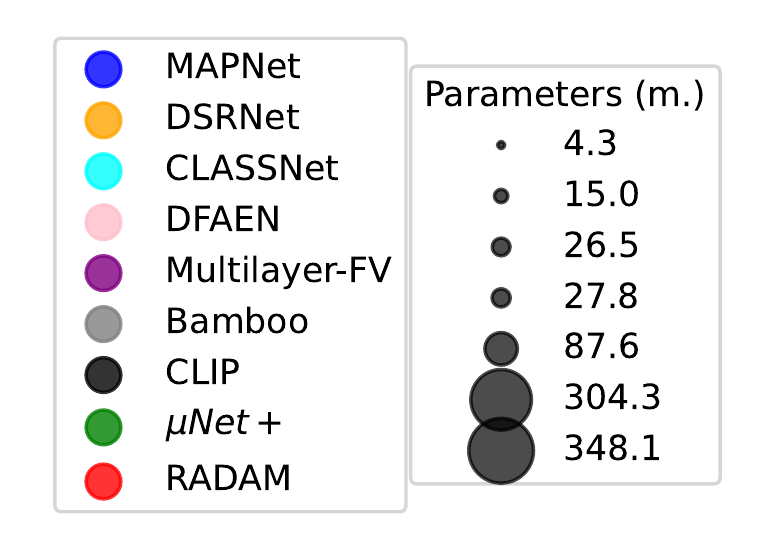}
  \subfigure[Average performance of methods with results available on the other four datasets.]{
   \centering
    \includegraphics[width=0.33\linewidth]{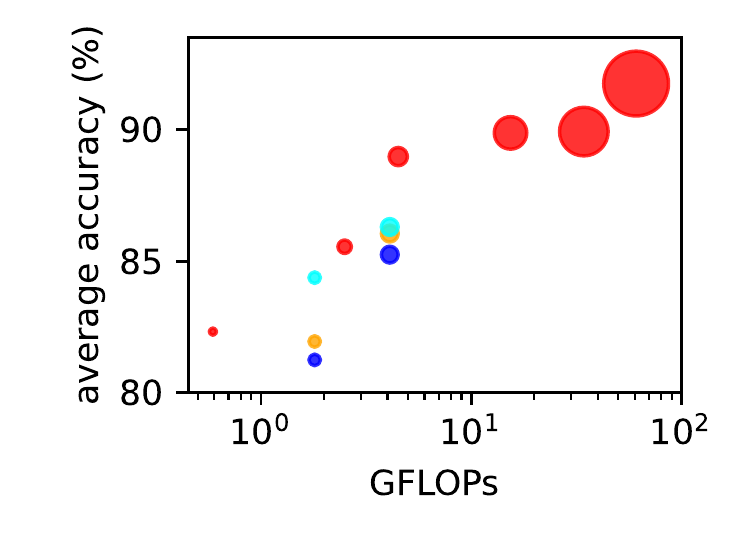}
    \label{fig:cost-avg}
    }
  \caption{Computational budget at inference time according to the backbone used by different methods.}
  \label{fig:cost}
\end{figure}

Additional analysis is performed to better understand the balance between the classification performance and computational budget of the compared texture recognition methods. We consider the inference costs in terms of GFLOPs and the number of parameters according to the backbone used by each method since this is the most resource-demanding step of every pipeline. One important aspect here is that the input size greatly impacts the FLOP count of the methods (check input sizes in parentheses in~\cref{tab:final-results}). Most works consider 224x224 inputs (the same input size employed by RADAM), and we assume this same size when not stated by the authors. For this analysis, we also are not considering the preparation of the backbone either in terms of pre-training cost or chosen dataset, nor the fine-tuning of the methods that do so. The results are shown in~\cref{fig:cost} first for the DTD dataset alone (\cref{fig:cost-dtd}) since this is the most challenging task and with more methods to compare, and then as the average for the remaining datasets (\cref{fig:cost-avg}). It is possible to notice the superiority of RADAM at different budgets, especially when using MobileNet V2 1.4 and ConvNeXt-T. Moreover, results also scale with backbone complexity. 

It is also important to mention the backbone costs of other methods not present in this analysis due to the lack of available results. For instance, Multilayer-FV achieved the previous state-of-the-art on FMD using EfficientNet-B5 with an input size of 512 pixels, which yields an approximate inference cost of 12 GFLOPs. This is considerably higher than the cost of ConvNeXt-T (4.5 GFLOPs), with which RADAM achieves an improvement of 4.3$\%$ in absolute performance compared to Multilayer-FV. On KTH, RankGP-3M-CNN++ achieves 91.1$\%$ accuracy (previous state-of-the-art) using three backbones, with a total inference cost of around 30 GFLOPs, while RADAM with ConvNeXt-T achieves comparable results ($-0.1\%$), and also achieves the state-of-the-art results using ConvNeXt-B (91.8$\%$ with inference cost of 15 GFLOPs), and ConvNeXt-XL (94.4$\%$ with inference cost of 61 GFLOPs). Considering GTOS and GTOS-Mobile, a competitive cost and performance are also achieved with RADAM using ConvNeXt-T, and state-of-the-art results with ConvNeXt-B, L, and XL.

To increment the cost analysis, we show in Table~\ref{tab:cost} the practical running time of RADAM (with $m=4$ and SVM) in contrast to the costs of ResNet. Our intuition is to compare the use of RADAM, which requires only a forward pass of the backbone and an SVM, to the approach in previous works of fine-tuning the backbone. We measured the average time (in milliseconds) of 100 runs of ResNet18 and ResNet50 over one 224x224 RGB image considering the RADAM module, the ResNet forward and backward passes, and the corresponding times when fine-tuning the backbone our using RADAM for feature extraction with SVM inference. It is important to notice that the cost of RADAM varies according to the backbone since the number of aggregate features ($z$) changes, thus impacting the RAE training cost. Considering the results in the table, fine-tuning ResNet50 on GTOS-Mobile using only 10 epochs with a batch size of 1 would take around 40 hours on CPU or 9 hours on GPU. On the other hand, extracting features with RADAM followed by SVM inference on the whole GTOS-Mobile dataset takes around 1.3 hours on CPU or half an hour on GPU. The training of SVM on the whole dataset (on CPU) takes an additional 15 minutes on average, without hyperparameter tuning. The results demonstrate that the RADAM module is considerably faster than the ResNet backbone, both in terms of inference speed and comparing feature extraction followed by SVM with training/fine-tuning. These results extend to ConvNeXt-nano and ConvNeXt-T, considering that the cost is comparable to ResNet18 and ResNet50, respectively, corroborating our claims that RADAM provides both considerable savings in training time and SOTA results at a similar inference cost. Moreover, considering more costly backbones such as ConvNext-L and XL, the gains in training time can be even more expressive.

\begin{table}
	\centering
	\caption{\label{tab:cost}Average running time (in milliseconds) of 100 repetitions using a 224x224 RGB image, performed on a machine with a GTX 1080ti, Intel Core i7-7820X 3.60GHz processor (using 8 threads), and 64GB of RAM.}
	%\resizebox{\linewidth}{!}{
	\begin{tabular}{cc|cc}
		\hline
		&&\multicolumn{2}{c}{time (ms)}\\
    	method & backbone & CPU & GPU\\
     	\hline \hline

        RADAM module alone &ResNet18& 2.7$\pm0.2$ & 2.9$\pm0.04$\\
        
        % RADAM module + SVM inference&ResNet18& 4.7$\pm0.5$ & 3.8$\pm0.03$\\
        
        backbone's forward pass &ResNet18&15.0$\pm1.6$& 4.3$\pm0.1$\\
        backbone's backward pass &ResNet18&50.8$\pm5.3$& 9.9$\pm1.5$\\

        backbone's 1 fine-tuning epoch&ResNet18& 65.8$\pm6.0$ & 14.2$\pm1.6$\\   
        backbone's forward pass + RADAM + SVM inference&ResNet18& 19.8$\pm1.6$ & 8.1$\pm0.1$\\
        
        \hline 
        RADAM module alone&ResNet50& 7.0$\pm0.4$ & 5.2$\pm0.4$\\

        % RADAM module + SVM inference&ResNet50& 16.2$\pm0.7$ & 7.0$\pm0.2$\\
        backbone's forward pass &ResNet50& 34.0$\pm2.6$ & 11.1$\pm0.8$ \\
        backbone's backward pass &ResNet50& 107.6$\pm5.9$ & 21.5$\pm2.2$ \\

        backbone's 1 fine-tuning epoch&ResNet50& 141.7$\pm7.6$ & 32.6$\pm2.3$\\
        backbone's forward pass + RADAM + SVM inference &ResNet50& 50.3$\pm2.6$ & 18.1$\pm0.8$\\

        % RADAM module&ResNet18& 2.9$\pm0.04$ & 2.7$\pm0.1$\\
        
        % backbone's forward pass &ResNet18& 4.5$\pm1.0$& 14.8$\pm1.8$\\
        % backbone's backward pass &ResNet18& 11.0$\pm2.4$& 49.8$\pm5.5$\\

        % backbone's 1 training epoch&ResNet18& 15.5$\pm3.0$ & 64.6$\pm6.9$\\

        % backbone + RADAM feature extraction&ResNet18& 7.4$\pm1.0$ & 17.5$\pm1.8$\\
        
        % \hline 
        % RADAM module&ResNet50& 7.3$\pm0.9$ & 7.5$\pm0.6$\\
        
        % backbone's forward pass &ResNet50& 11.2$\pm1.4$ & 35.1$\pm4.4$ \\
        % backbone's backward pass &ResNet50& 22.4$\pm2.2$ & 112.6$\pm9.4$ \\

        % backbone's 1 training epoch&ResNet50& 33.6$\pm3.1$ & 147.7$\pm12.3$\\
        % backbone + RADAM feature extraction&ResNet50& 18.5$\pm1.6$ & 42.6$\pm4.9$\\

    \hline

	\end{tabular}
	%}
\end{table}

\section{Conclusion}
We presented RADAM, a new feature encoding module for texture analysis. The method consists of randomly encoding aggregated deep activation maps from pre-trained DCNN using RAEs. These autoencoders learn to pool activation maps into a 1-dimensional representation by training on its $z$-dimentional pixels as sample points. A texture image is then encoded by using the decoder weights learned from its activation maps. The procedure is orderless, but takes into account the spatial information of the pixels by using a 2D positional encoding. Compared to previous works, our method does not require fine-tuning of the backbone, and the encoding module is rather simple. Linear classification of the descriptors is performed with an SVM, and we achieve state-of-the-art performance on several texture benchmarks. RADAM also achieves the best efficiency considering inference cost and performance using backbones with varying computational budgets. These results are impressive also considering that, compared to other methods, no fine-tuning of the backbone is needed for RADAM, causing a lower cost also at training time.
%These results suggest that the pre-trained backbones contain rich learned representations that can be explored without the cost of fine-tuning them specifically for texture recognition tasks.

Our work corroborates a simpler approach to texture recognition where the fine-tuning of costly backbones may not be necessary to achieve high discriminatory power. For future works, one may explore different backbones or different formulations of our RAE, with multiple layers, more hidden neurons, and other possible improvements. On the other hand, if enough computing resources are available, another approach more similar to previous works would be to explore our module in an end-to-end manner. Since RADAM is deterministic and a closed-form solution, an alternative would be adding a linear layer instead of an SVM and optimizing it along with the backbone.

\section*{Acknowledgements}

L. Scabini acknowledges funding from the São Paulo Research Foundation (FAPESP) (Grants \#2019/07811-0 and \#2021/09163-6) and the National Council for Scientific and Technological Development (CNPq) (Grant \#142438/2018-9). K M. Zielinski acknowledges support from São Paulo Research Foundation (FAPESP) (Grant \#2022/03668-1) and Higher Education Personnel Improvement Coordination (CAPES) (Grant  \#88887.631085/2021-00).  L. C. Ribas acknowledges support from São Paulo Research Foundation (FAPESP) (grants \#2021/07289-2 and \#2018/22214-6). W. N. Gon\c{c}alves acknowledges support from CNPq (Grants \#405997/2021-3 and \#305296/2022-1). O. M. Bruno acknowledges support from CNPq (Grant \#307897/2018-4) and FAPESP (grants \#2018/22214-6 and \#2021/08325-2). The authors are also grateful to the NVIDIA GPU Grant Program. This research received funding from the Flemish Government under the "Onderzoeksprogramma Artificiële Intelligentie (AI) Vlaanderen" programme.

%%%%%%%%% REFERENCES
{\small
\bibliographystyle{ieee_fullname}
\bibliography{arxiv}
}



\section*{Supplementary material}

# Supplementary Material for RADAM: Texture Recognition through Randomized Aggregated Encoding of Deep Activation Maps

## 1 RADAM implementation

We give the exact python code for the RADAM module and all its internal mechanisms. In the future, we will make this implementation and the experiments available in a public repository. First of all, RADAM is applied at the output of a backbone built using the *timm* library. We do that using the following call:

```python
import timm
net = timm.create_model(model, features_only=True, pretrained=True, output_stride=8)
```

Here, *model* is a string with the name of the backbone according to the library (*e.g.* resnet18, convnext_nano, etc). The *features only* option works for most convolutional architectures and turns the output of the model into a list of activation maps at different depths. In this sense, the following operations can be used to get some properties from these activation maps:

```python
import numpy as np
import torch
device = "cuda" if torch.cuda.is_available() else "cpu"
net.to(device)
act_maps = net(torch.zeros(1,3,224,224))
z = sum([d.size()[1] for d in act_maps])
half_depth = int(np.round(len(act_maps)/2))
w = act_maps[half_depth].size()[2]
h = act_maps[half_depth].size()[3]
```

These properties are, respectively, the sum of the number of features of each activation map $z$, and the spatial dimensions at approximately the middle of the architecture (w,h). Notice that these properties can be computed for any input image size aside from (224,224) if accepted by the backbone. They are then used to create the main class for RADAM:

```python
import torchvision
from einops.layers.torch import Rearrange
class RADAM(torch.nn.Module):
    def __init__(self, device, z, spatial_dims = (28,28), m=4, pos_encoding=True,
        dim_norm=(2,3)):
        super(RADAM, self).__init__()
        self.q=1
        self.z=z
        self.device = device
        self.rearange = torch.nn.Sequential(lp_norm_layer(p=2.0, dim=dim_norm, eps=1e-10),
                                torchvision.transforms.Resize(spatial_dims),
                                Rearrange('b c h w -> b c (h w)')
                                )
        self.RAEs = []
        for model in range(m):
            self.RAEs.append(RAE(q=self.q, z=z, w=spatial_dims[0], h=spatial_dims[1],
                device=device, seed=model*(self.q*z), pos_encoding=pos_encoding))
```

```
16
17      def forward(self, x):
18          out = []
19          if not isinstance(x, list):
20              x = [x]
21          device = self.device
22          batch, _, _, _ = x[0].size()
23          for depth in range(len(x)):
24              x[depth] = self.rearange(x[depth])
25          for sample in range(batch):
26              agg_activation_map = torch.vstack([x[i][sample] for i in range(len(x))])
27              pooled_features = torch.zeros(self.q,self.z).to(device)
28              for rae in self.RAEs:
29                  pooled_features += rae.fit_AE(agg_activation_map)
30              out.append(pooled_features)
31          return torch.stack(out)
32
33  class lp_norm_layer(torch.nn.Module):
34      def __init__(self,p=2.0, dim=(1,2), eps=1e-10):
35          super().__init__()
36          self.p=p
37          self.dim=dim
38          self.eps=eps
39      def forward(self, x):
40          return torch.nn.functional.normalize(x, p=self.p, dim=self.dim, eps=self.eps)
```

The parameters in the code are the ones used by our experiments in the main paper. Additionally, the *einops* library is used for efficient reorganization of the activation map, and we also use an additional class to perform $L_p$ normalization in batches as a PyTorch layer.

The implementation of the Randomized Autoencoder (RAE) is shown in the following. It takes as arguments the dimensions of the model, previously passed to the RADAM class. The constructor generates the random encoder and the positional encoding, and the *fit_AE* method computes the decoder weights, i.e, the least-squares training:

```
1   class RAE():
2       def __init__(self, q, z, w, h, device, pos_encoding, seed):
3           self._device = device
4           self.pos_encoding=pos_encoding
5           self.W = make_orthogonal(LCG(q, z, seed)).to(device)
6           if self.pos_encoding:
7               self.encoding = positionalencoding2d(z, w, h)
8               self.encoding = torch.reshape(self.encoding, (z, w*h)).to(device)
9           self._activation = torch.sigmoid
10
11      def fit_AE(self, x):
12          if self.pos_encoding:
13              x = torch.add(x, self.encoding)
14          g = self._activation(torch.mm(self.W, x))
15          return torch.linalg.lstsq(g.t(), x.t()).solution
```

The remaining functions used for weight initialization and positional encoding are given in the following:

```
1   import pickle
2   def LCG(m, n, seed):
3       L = m*n
4       # To generate this weight file, use:
5           #V = torch.zeros(2**18, dtype=torch.float)
6           #V[0], a, b, c = 1, 75, 74, (2**16)+1
7           #for x in range(1, (2**18)):
8           #    V[x] = (a*V[x-1]+b) % c
9           #with open('RAE_LCG_weights.pkl', 'wb') as f:
10          #pickle.dump(V, f)
11      with open('RAE_LCG_weights.pkl', 'rb') as f:
```

```
12          V = pickle.load(f)
13          f.close()
14      V = V[seed:L+seed]
15      V = torch.divide(torch.subtract(V, torch.mean(V)), torch.std(V))
16      return V.reshape((m,n))
17
18  def make_orthogonal(tensor):
19      rows = tensor.size(0)
20      cols = tensor.numel() // rows
21      flattened = torch.reshape(tensor, (rows, cols))
22      if rows < cols:
23          flattened.t_()
24      q, r = torch.linalg.qr(flattened)
25      d = torch.diag(r, 0)
26      ph = d.sign()
27      q *= ph
28      if rows < cols:
29          q.t_()
30      return q
31
32  import math
33  def positionalencoding2d(d_model, height, width):
34      pe = torch.zeros(d_model, height, width)
35      d_model = int(d_model / 2)
36      div_term = torch.exp(torch.arange(0., d_model, 2) *
37                           -(math.log(10000.0) / d_model))
38      pos_w = torch.arange(0., width).unsqueeze(1)
39      pos_h = torch.arange(0., height).unsqueeze(1)
40      pe[0:d_model:2, :, :] = torch.sin(pos_w * div_term).transpose(0, 1).unsqueeze(1).repeat(1,
        ↪ height, 1)
41      pe[1:d_model:2, :, :] = torch.cos(pos_w * div_term).transpose(0, 1).unsqueeze(1).repeat(1,
        ↪ height, 1)
42      pe[d_model::2, :, :] = torch.sin(pos_h * div_term).transpose(0, 1).unsqueeze(2).repeat(1,
        ↪ 1, width)
43      pe[d_model + 1::2, :, :] = torch.cos(pos_h * div_term).transpose(0,
        ↪ 1).unsqueeze(2).repeat(1, 1, width)
44      return pe
```

Finally, the main *RADAM* class can be simply applied to the output of *net*, for a given image or batch, to obtain texture representations:

```
1  texture_representation = RADAM(device, z, (w,h))(net(input_batch))
```

As detailed in the main paper, these representations are then used to train a linear SVM for texture classification. For that, we use the scikit-learn library to build the classifier as follows:

```
1  from sklearn import svm
2  SVM = svm.SVC(kernel='linear')
```

## 2 Additional Experiments with RADAM

Here we present additional experiments not present in the main paper.

### 2.1 Variance caused by random initializer

Tab. 1 shows the RADAM performance variance caused by changing the LCG configuration ($a$, $b$, and $c$) for weight initialization of the RAE. We selected a set of 10 configurations from the table available in [1]. In general, we observe minimal performance variance concerning these parameters, especially with larger backbones, corroborating that the choice of the random weights is not critical. We adopted parameters $a = 75$, $b = 74$, and $c = 2^{16} + 1$ starting with $x = 1$ (first line in the table), as it uses the smallest integers and is also a classical configuration according to the ZX81

---

[1] www.wikipedia.org/wiki/Linear_congruential_generator

computer from 1981. However, small variations may impact the comparison between RADAM and other methods. Nevertheless, it is important to notice that our choice of LCG parameters does not impact the new state-of-the-art results reported in the main paper for RADAM. In fact, carefully selecting specific configurations in Tab. 1 for each dataset and backbone yields even higher results, but we preferred a single consistent configuration to be used for all cases. Additionally, for applicability purposes, we provide files with the exact values computed by our code for all the LCG configurations (which will be available in our code repository).

Table 1: The variance on classification accuracy of RADAM when changing the LCG configuration to generate random weights for the encoder. For the second part of the table (ConvNeXt results), the LCG configurations correspond to the same rows as for ResNet18.

| | Parameters of LCG | | | ResNet18 | | | |
|---|---|---|---|---|---|---|---|
| source | a | b | c | DTD | FMD | KTH | Outex13 |
| ZX81 | 75 | 74 | $2^{16}+1$ | 68.1 | 77.7 | 84.7 | 89.0 |
| Numerical Recipes book | 1664525 | 1013904223 | $2^{32}$ | 68.0 | 77.9 | 85.0 | 90.0 |
| Borland C/C++ | 22695477 | 1 | $2^{32}$ | 66.7 | 79.7 | 86.0 | 88.2 |
| glibc (used by GCC) | 1103515245 | 12345 | $2^{31}$ | 68.0 | 78.0 | 84.7 | 90.0 |
| Borland Delphi, Virtual Pascal | 134775813 | 1 | $2^{32}$ | 67.9 | 79.2 | 85.6 | 90.1 |
| Microsoft Visual/Quick C/C++ | 214013 | 2531011 | $2^{32}$ | 68.1 | 78.2 | 85.1 | 89.3 |
| Microsoft Visual Basic | 1140671485 | 12820163 | $2^{24}$ | 68.1 | 79.2 | 85.4 | 88.7 |
| RtlUniform from Native API | 2147483629 | 2147483587 | $2^{31}-1$ | 68.0 | 78.9 | 85.1 | 89.6 |
| Java's java.util.Random | 25214903917 | 11 | $2^{48}$ | 67.9 | 77.4 | 85.4 | 89.7 |
| cc65 | 16843009 | 826366247 | $2^{32}$ | 68.0 | 78.7 | 85.6 | 89.0 |
| | | | average | 67.9±0.4 | 78.5±0.8 | 85.3±0.4 | 89.4±0.6 |

| | ConvNeXt-nano | | | | ConvNeXt-XL in ImageNet-21K | | | |
|---|---|---|---|---|---|---|---|---|
| | DTD | FMD | KTH | Outex13 | DTD | FMD | KTH | Outex13 |
| | 74.9 | 87.1 | 89.6 | 92.5 | 83.7 | 95.2 | 94.4 | 92.5 |
| | 75.0 | 86.9 | 89.2 | 92.4 | 83.9 | 95.7 | 94.5 | 91.8 |
| | 75.1 | 86.8 | 89.5 | 91.2 | 83.7 | 95.3 | 94.6 | 92.4 |
| | 75.0 | 86.6 | 89.8 | 92.2 | 83.9 | 95.4 | 94.1 | 92.6 |
| | 75.1 | 86.4 | 89.5 | 92.4 | 83.8 | 95.6 | 94.5 | 92.4 |
| | 75.1 | 87.1 | 89.5 | 92.5 | 83.8 | 95.5 | 94.6 | 92.8 |
| | 74.9 | 87.7 | 89.4 | 92.2 | 83.7 | 95.3 | 94.6 | 92.4 |
| | 75.1 | 86.9 | 89.6 | 91.9 | 83.8 | 95.2 | 94.6 | 91.9 |
| | 74.9 | 86.9 | 89.3 | 92.6 | 83.8 | 95.5 | 94.8 | 92.2 |
| | 75.0 | 87.6 | 89.0 | 92.2 | 83.8 | 95.5 | 94.5 | 92.8 |
| average | 75.0±0.1 | 87.0±0.4 | 89.4±0.2 | 92.2±0.4 | 83.8±0.1 | 95.4±0.2 | 94.5±0.2 | 92.4±0.3 |



\end{document}